\documentclass[a4paper]{cas-dc}
\usepackage[numbers]{natbib}

\def\tsc#1{\csdef{#1}{\textsc{\lowercase{#1}}\xspace}}
\tsc{WGM}
\tsc{QE}
\tsc{EP}
\tsc{PMS}
\tsc{BEC}
\tsc{DE}

\begin{document}
\let\WriteBookmarks\relax
\def\floatpagepagefraction{1}
\def\textpagefraction{.001}
\shorttitle{}
\shortauthors{Z. Xia et~al.}

\title [mode = title]{TCSAFormer: Efficient Vision Transformer with Token Compression and Sparse Attention for Medical Image Segmentation}

\author{Zunhui Xia}[style=chinese,orcid=0009-0008-6706-5817]
\ead{zunhui.xia@stu.cqut.edu.cn}

\credit{Data curation, Writing - Original draft preparation}

\author{Hongxing Li}[style=chinese]

\author{Libin Lan}[
                        style=chinese,
                        orcid=0000-0003-4754-813X
                        ]
\cormark[1]
\ead{lanlbn@cqut.edu.cn}

\credit{Conceptualization of this study, Methodology, Software}

\affiliation{organization={College of Computer Science and Engineering, Chongqing University of Technology},
                city={Chongqing},
              citysep={}, 
                postcode={400054}, 
                state={Chongqing},
                country={China}}

\cortext[cor1]{Corresponding author}

\begin{abstract}
In recent years, transformer-based methods have achieved remarkable progress in medical image segmentation due to their superior ability to capture long-range dependencies. However, these methods typically suffer from two major limitations. First, their computational complexity scales quadratically with the input sequences. Second, the feed-forward network (FFN) modules in vanilla Transformers typically rely on fully connected layers, which limits models’ ability to capture local contextual information and multiscale features critical for precise semantic segmentation. To address these issues, we propose an efficient medical image segmentation network, named TCSAFormer. The proposed TCSAFormer adopts two key ideas. First, it incorporates a Compressed Attention (CA) module, which combines token compression and pixel-level sparse attention to dynamically focus on the most relevant key-value pairs for each query. This is achieved by pruning globally irrelevant tokens and merging redundant ones, significantly reducing computational complexity while enhancing the model's ability to capture relationships between tokens. Second, it introduces a Dual-Branch Feed-Forward Network (DBFFN) module as a replacement for the standard FFN to capture local contextual features and multiscale information, thereby strengthening the model’s feature representation capability. We conduct extensive experiments on three publicly available medical image segmentation datasets: ISIC-2018, CVC-ClinicDB, and Synapse, to evaluate the segmentation performance of TCSAFormer. Experimental results demonstrate that TCSAFormer achieves superior performance compared to existing state-of-the-art (SOTA) methods, while maintaining lower computational overhead, thus achieving an optimal trade-off between efficiency and accuracy. The code is available on \href{https://github.com/XiaZunhui/TCSAFormer}{GitHub}.
\end{abstract}

\begin{keywords}
Efficient Vision Transformer, \sep Medical image segmentation,  \sep Sparse attention, \sep TCSAFormer, \sep Token compression.
\end{keywords}

\maketitle

\section{Introduction}
Medical image segmentation enables precise pixel-level classification of target anatomical structures or lesion regions, providing quantitative analysis for clinical decision-making. 
However, conventional medical image segmentation methods exhibit limited segmentation performance when confronted with complex anatomical structures and ambiguous boundaries \cite{liang2023maxformer,azad2024advances,xu2023dcsau}. Therefore, developing more precise medical image segmentation techniques to address these challenges is of great significance for enhancing clinical diagnosis and treatment.

In recent years, convolutional neural network (CNN)-based  encoder-decoder architectures such as U-Net \cite{ronneberger2015unet}, UNet++ \cite{zhou2018unet++}, and UNet 3+ \cite{huang2020unet3+} , achieve significant progress in medical image segmentation. These methods leverage local receptive fields to capture fine-grained local features, demonstrating outstanding performance in boundary-sensitive segmentation tasks. However, constrained by the inherent locality of convolution operations, aforementioned approaches exhibit limitations in modeling long-range dependencies. Although some efforts attempt to mitigate this issue by introducing dilated convolutions \cite{gu2019cenet,feng2020cpfnet} or pyramid pooling \cite{zhao2017pyramid}, their global context modeling capability remains significantly inadequate, especially in tasks requiring comprehensive semantic understanding, such as multi-organ segmentation.
\begin{figure}
\centering
\includegraphics[width=1\linewidth]{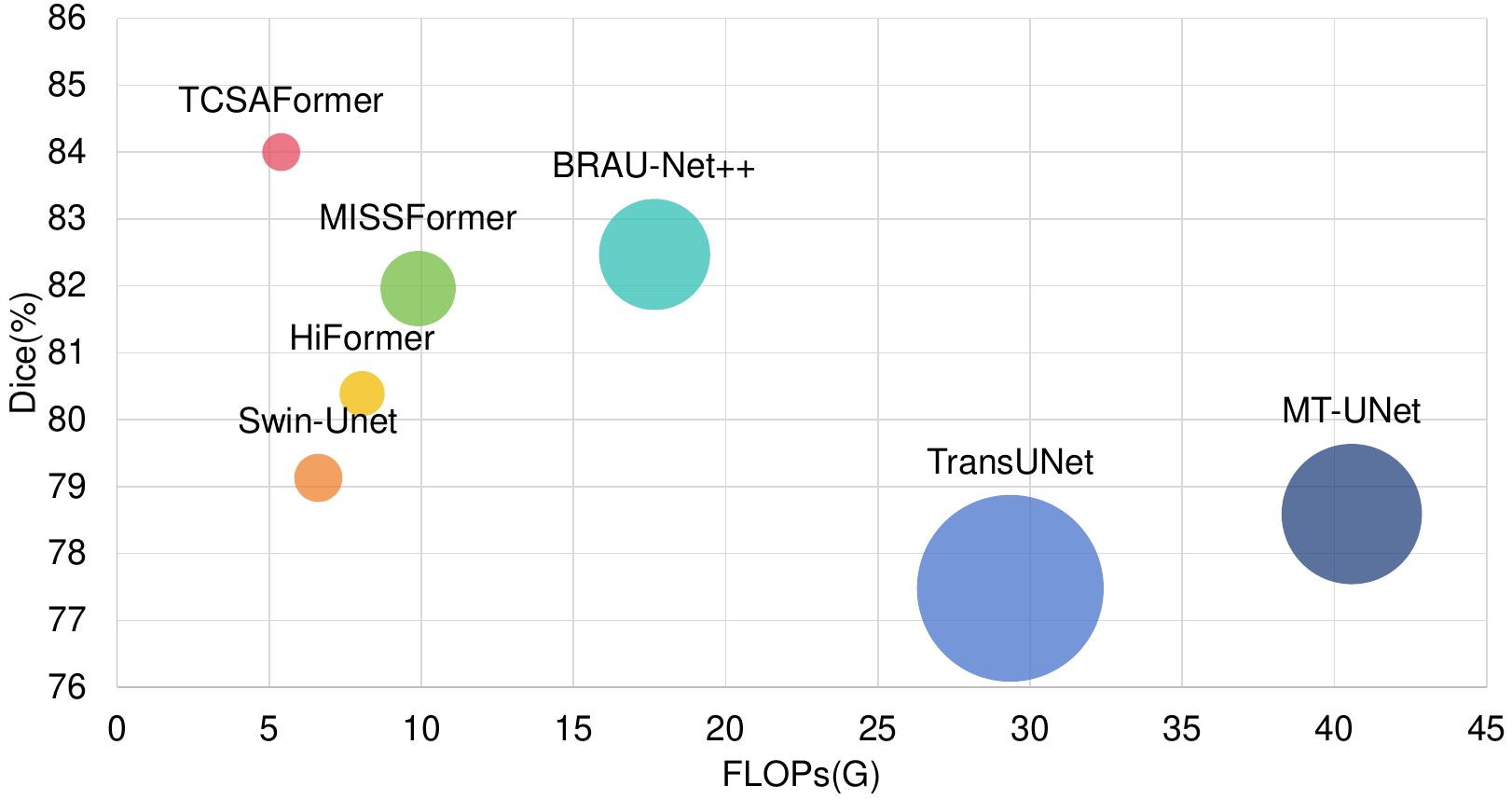}
\caption{Performance comparison of different medical image segmentation models on the Synapse dataset. The x-axis represents FLOPs, the y-axis denotes Dice scores, and the circle size indicates parameters. Our proposed TCSAFormer achieves optimal segmentation accuracy with fewer parameters and lower computational cost, demonstrating an effective trade-off between efficiency and accuracy.}
\label{motivateion}
\end{figure}

Vision Transformer architectures leverage self-attention mechanisms effectively capture global contextual information. Building on this advantage, CNN-Transformer hybrid structures \cite{chen2021transunet,zhang2021transfuse} and pure transformer \cite{cao2022swin} architectures 
 are proposed for medical image segmentation, effectively extracting cross-region contextual information such as correlations among multiple organs and continuous boundaries. However, vanilla transformer adopted by models like TransUNet \cite{chen2021transunet} and TransFuse \cite{zhang2021transfuse} exhibit a potential drawback: the computational complexity of standard self-attention mechanisms grows quadratically with input image resolution, resulting in substantial computational burden and memory footprint. To address this challenge, the prior works develop various sparse attention mechanisms to construct medical segmentation networks. Early studies primarily focused on sparse patterns with spatial structures. The core idea behind these methods is to design predefined fixed patterns, which limit queries to attend key-value pairs within a specified range, without requiring additional complex operations. Representative works include MedT \cite{valanarasu2021medt}, SwinUnet \cite{cao2022swin}, and HiFormer \cite{heidari2023hiformer}. Nevertheless, these handcrafted sparse patterns are query-agnostic, they cannot adaptively handle varying data. To overcome this issue, subsequent research shifts toward content-aware sparse attentions. BRAU-Net \cite{Cai2025BRAU-Net} and BRAU-Net++ \cite{Lan2024BRAU-Net++} employ bi-level routing attention mechanism \cite{zhu2023biformer} to dynamically select highly relevant key-value pairs for queries, largely preserving global modeling capability. Despite these advances, two critical issues persist: First, sparse selection at the coarse-grained regional level overlooks token-specific requirements within regions, making it difficult to model inter-token relationships. Second, globally low-relevance tokens such as background and non-target regions still participate in attention computations, which increases unnecessary computational overhead while leading the model to learn irrelevant features, ultimately compromising overall performance.
Additionally, some transformer blocks still follow the conventional transformer design, using a two-layer MLP as the feed-forward network. This design is essentially a fully connected mapping that lacks the ability to model spatial relationships, making it unsuitable for semantic segmentation tasks requiring strong spatial understanding. Previous works demonstrate that introducing local context modeling and multiscale feature extraction is beneficial for enhancing the model's capacity to represent spatial information \cite{huang2022missformer, lin2022ds-transunet,srivastava2021msrf,shen2025smanet}. Therefore, in this work, we consider redesigning the FFN by incorporating local contextual information and multiscale features to improve the segmentation performance in complex scenes.

To address these challenges, we propose TCSAFormer, an efficient medical image segmentation network integrating a Compression Attention (CA) module  for computational burden reduction and a Dual-Branch Feed-Forward Network (DBFFN) that enhances feature representation capability. First, we construct CA by combining a Token Compression Pipeline (TCP) \cite{rao2021dynamicvit,bolya2023tome,tang2022patch,norouzi2024algm} with a content-aware Top-k Sparse Sttention (TKSA) mechanism \cite{zhao2019explicit, wang2022kvt} to reduce the computational complexity of transformers. Specifically, TCP employs a lightweight predictor to evaluate token importance, subsequently performing token pruning and similarity-driven token merging to effectively reduce computational load while preserving critical information. Concurrently, TKSA breaks through the limitations of conventional window-based sparsity by dynamically selecting high relevance tokens for each token at the pixel level. Notably, we design a Token Decompression Pipeline (TDP) within CA to reconstruct spatial topology after attention computation, avoiding the loss of important information. Second, to enhance the feature representation capability, we design DBFFN, which utilizes a dual-branch convolutional structure to simultaneously capture local context and multiscale features. Based on the above strategy, TCSAFormer achieves a trade-off between accuracy and efficiency. Fig.\ref{motivateion} compares the segmentation performance and computation complexity of TCSAFormer with those of state-of-the-art medical image segmentation methods.

Overall, our contributions can be summarized as follows:
\begin{enumerate}
    \item We propose CA, which combines token compression with a sparse attention mechanism to reduce model complexity while focusing on crucial features.
    \item We design DBFFN that utilizes receptive fields at multiple scales to capture both coarse-grained regional features and fine-grained local spacial details, thereby improving the feature representation capability in complex scenarios.
    \item Based on CA and DBFFN, we propose TCSAFormer, a novel encoder-decoder symmetric network for medical image segmentation. We evaluate TCSAFormer on three publicly available datasets, and the experimental results demonstrate its superior performance in medical image segmentation tasks.
\end{enumerate}

The remainder of the paper is organized as follows. Section \ref{releted work} reviews the related work and analyzes the shortcomings of existing methods. Section \ref{methods} provides a detailed description of the proposed approach, including the overall architecture of TCSAFormer and specific modules such as CA and DBFFN. Section \ref{experiment} presents the experimental settings and results. Section \ref{conclusion} summarizes the main contributions, existing limitations, and future research directions.

\section{Related Work}
\label{releted work}
\subsection{Token Compression in Vision Transformer}
\label{token compression}
The purpose of token compression is to reduce the number of tokens involved in self-attention calculations, thereby decreasing computational complexity. Current works mainly follow two technical paths: token pruning and token merging.
Token pruning reduces the computation by dynamically estimating the importance of each token and removing unimportant tokens layer by layer, such as DynamicViT \cite{rao2021dynamicvit} and AdaViT \cite{meng2022adavit}. However, these methods have a common issue: the pruned tokens no longer participate in subsequent layer calculations, potentially leading to the loss of useful information and degrading model performance. On the other hand, ToMe \cite{bolya2023tome} and PITOME \cite{tran2024pitome} avoid redundant computation by merging similar tokens rather than directly pruning them, which can effectively alleviate the issue of information loss. While this method effectively addresses the issue of information loss,  completely removing token pruning may compromise the advantages of discarding insignificant tokens and impair the model's generalization ability.

Based on the advantages and drawbacks of these two techniques, we propose a new token compression pipeline that adopts a "prune first, merge later" strategy. This approach can discard insignificant tokens while merging similar tokens to avoid redundant computation. In addition, to prevent the potential loss of information during compression, we design a token decompression pipeline that restores the compressed tokens to their original spatial positions.

\subsection{Sparse Attention in Medical Image Segmentation}
In recent years, Transformer-based medical image segmentation models have effectively alleviated computational complexity issues through the introduction of sparse attention mechanisms. Early approaches adopt handcrafted sparse patterns that reduce computational costs by restricting the range of attention for queries. For instance, MedT \cite{valanarasu2021medt} and MSAANet \cite{zeng2023msaanet} restricts attention along both height and width dimensions, SwinUnet \cite{cao2022swin} and HiFormer \cite{heidari2023hiformer} confine attention within local windows, while MT-UNet \cite{wang2022mt-unet} integrates both strategies to enhance global modeling capabilities. However, these handcrafted sparse patterns suffer from a critical limitation: different queries share identical key-value pairs, thereby restricting dynamic feature interaction capabilities. To address this issue, BRA-UNet++ \cite{Lan2024BRAU-Net++} achieves dynamic token selection via a bi-level routing attention. Unlike these methods, we propose a sparse attention based on a learnable Top-k operation, which enables precise modeling of token relationships for pixel-level token selection. Furthermore, we integrate a token compression method with the sparse attention mechanism to effectively filter globally insignificant tokens, thereby reducing unnecessary computational overhead, this is a critical aspect overlooked by existing sparse approaches.
\subsection{Feed-Forward Networks in Vision Transformer}
The standard Transformer model employs two fully connected layers as the feed-forward network (FFN) to integrate information. However, this approach fails to effectively capture local contextual information. To address this limitation, SegFormer \cite{xie2021segformer} and PVT v2 \cite{wang2022pvtv2} introduce a 3$\times$3 depthwise convolution in FFN to enhance local continuity. Building on this, GhostViT \cite{cao2023ghostvit} design the cheap Ghost FFN module to generate more diverse features by performing light-weight expand–reduce transformation. In the field of medical image segmentation, MISSFormer \cite{huang2022missformer} and DMSA-UNet \cite{li2024dmsa} respectively proposed the ReMix-FFN module for feature alignment and the CGLU module for adaptively enhancing or suppressing features. In this work, we continue to employ depthwise convolutions to capture local contextual information, but with a key innovation: we propose a dual-branch structure to facilitate feature interactions at multiple scales, thereby enhancing the model's feature representation capabilities.

\begin{figure*}
\centering
\includegraphics[width=\textwidth]{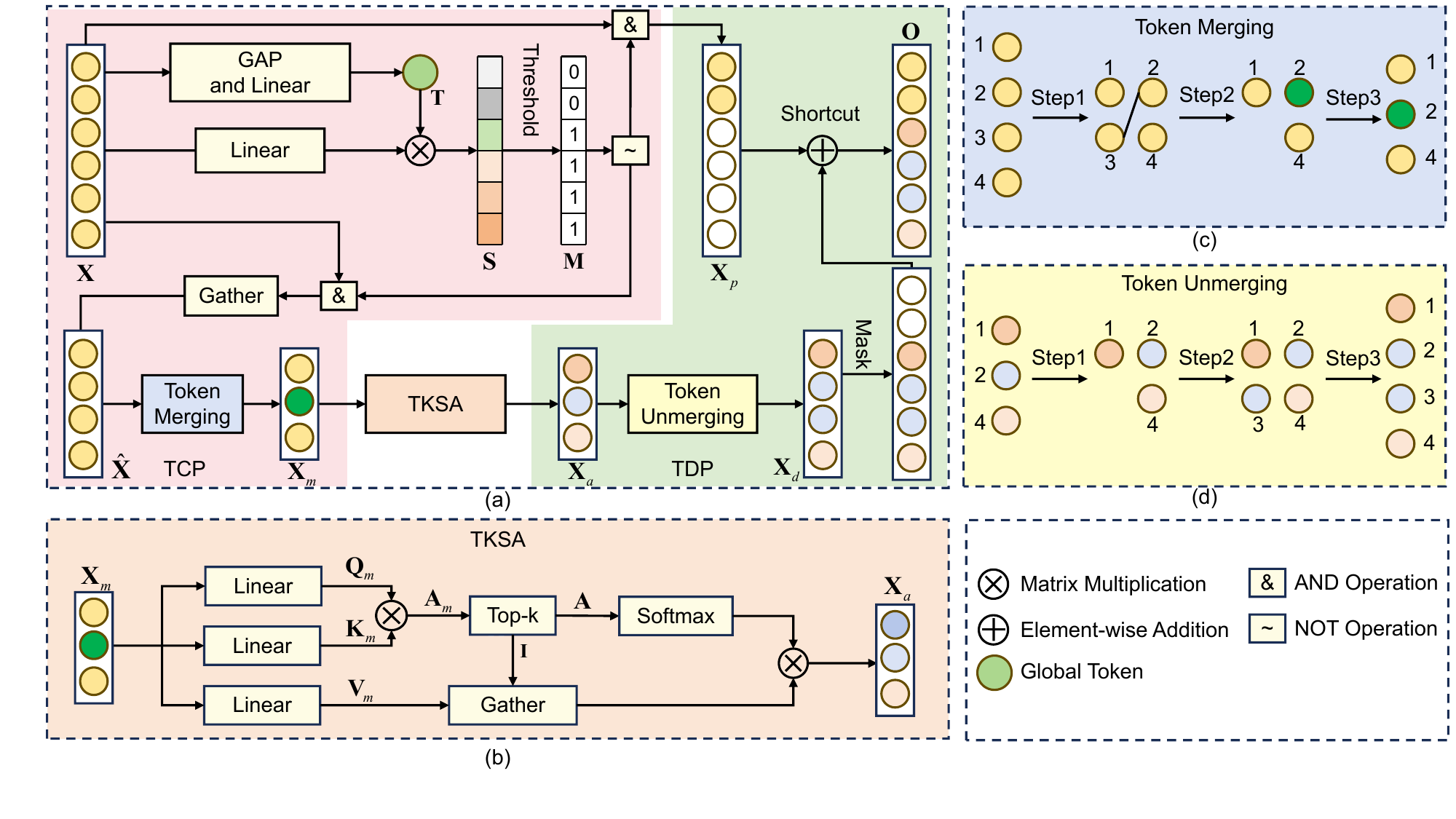}
\caption{(a) illustrates the overall architecture of CA, where the {\color[rgb]{0.96,0.56,0.66}pink} region represents the token compression pipeline (TCP) and the \textcolor{green}{green} region represents the token decompression pipeline (TDP). (b) shows the computation process of Top-k sparse attention. (c) and (d) depict the steps of token merging and token unmerging, respectively.}
\label{CAM}
\end{figure*}

\section{Methods}
\label{methods}
In this section, we will sequentially introduce the components of TCSAFormer. We begin by providing a detailed explanation of the CA, which serves as one of the core components of TCSAFormer. Next, we will delve into the design details of the DBFFN. Building upon these, we will explain how the TCSAFormer block is constructed. Finally, we will present the overall architecture of TCSAFormer.
\subsection{Compression Attention}
Self-attention mechanisms offer benefits in modeling global context, but their computational complexity grows quadratically with the input sequence length. To tackle this issue, we propose CA module. The overall architecture of CA is illustrated in Fig. \ref{CAM}. Specifically, CA consists of three core components: a Token Compression Pipeline (TCP) designed to reduce the number of tokens; a Top-k sparse attention mechanism for modeling pixel-level long-range dependencies; and a Token Decompression Pipeline (TDP) tasked with restoring the spatial position information of the compressed tokens. In the subsequent sections, we provide implementation details for each component.
\subsubsection{Token Compression Pipeline}
As mentioned in Section \ref{token compression}, using token pruning or token merging individually has certain limitations. Therefore, we propose a dynamic Token Compression Pipeline that combines both methods, aiming to alleviate the computational burden while preserving critical information. 

The pipeline initially incorporates an adaptive token pruning module, which is guided by global semantic information to eliminate globally irrelevant tokens. Specifically, for the input feature tensor 
$\mathbf{X} \in \mathbb{R}^{HW \times C}$, spatial dimensions are compressed using global average pooling to capture global features. Then, a learnable linear projection matrix $\mathbf{W}_1 \in \mathbb{R}^{C \times C}$ is applied to generate a global semantic token $\mathbf{T} \in \mathbb{R}^{1 \times C}$ with contextual semantic information. 
\begin{equation}
    \mathbf{T}=(\mathrm{GAP}(\mathbf{X}))\mathbf{W}_1.
\end{equation}
After that, another projection matrix $\mathbf{W}_2 \in \mathbb{R}^{C \times C}$ projects the local tokens in $\mathbf{X}$ to the same semantic space as $\mathbf{T}$, allowing for the computation of global importance scores $\mathbf{S} \in \mathbb{R}^{HW \times 1}$ for each local token.
\begin{equation}
    \mathbf{S}=(\mathbf{X}\mathbf{W}_2)\mathbf{T}^T.
\end{equation}
Based on a predefined pruning rate $\rho$, the pruning threshold $\tau$ is dynamically determined, and a binary mask matrix $\mathbf{M}\in \mathbb{R}^{HW \times 1}$ for the local tokens is generated.
\begin{equation}
 \mathbf{M}_{i}=
\begin{cases}
1 & \mathbf{S}_{i} > \tau \\
0 & \mathbf{S}_{i} \leq \tau
\end{cases}, \quad \text{for } i=1,2,\dots,HW.
\end{equation}
Finally, the pruned token set $\mathbf{X}_p\in \mathbb{R}^{HW \times C}$ is obtained by applying simple logical NOT and AND operations.
\begin{equation}
    \mathbf{X}_p=\mathbf{X} \& \overline{\mathbf{M}}.
\end{equation}

Furthermore, we integrate a dynamic token merging into the pipeline. The token merging based on a bipartite graph to adaptively merge similar tokens without the additional learnable parameters. Before merging, we employ a simple logical AND operation to extract crucial tokens, then gather them into a new tensor to facilitate parallel operations, resulting in $\hat{\mathbf{X}} \in \mathbb{R}^{n \times C}$, $n$ is the number of tokens that are retained.
\begin{equation}
    n=HW*\rho,
\end{equation}
\begin{equation}
        \hat{\mathbf{X}}=\mathrm{Gather}(\mathbf{X} \& \mathbf{M}).
\end{equation}
The implementation of token merging consists of the following three steps: 
\begin{enumerate}
    \item \textbf{Bipartite Graph Construction}: The crucial tokens are evenly split into sets $\mathbf{A}\in \mathbb{R}^{\frac{n}{2} \times C}$ and $\mathbf{B}\in \mathbb{R}^{\frac{n}{2} \times C}$. Based on similarity, for each token in $\mathbf{A}$, an edge is drawn to the most similar token in set $\mathbf{B}$. Afterward, the top $r$ edges with the highest weights are retained through global sorting, where $r$ is dynamically determined by the merging rate $\rho_m$.
    \item \textbf{Feature fusion}: The left node connected by each edge in the bipartite graph is fused into the corresponding right node.
    \item \textbf{Set concatenation}: Set $\mathbf{B}$ is concatenated to the end of set $\mathbf{A}$ along the channel dimension.
\end{enumerate}
This procedure is illustrated in Fig. \ref{CAM} (c), and can be formulated as:
\begin{equation}
    r=n-(\rho_m*n),
\end{equation}
\begin{equation}
\mathbf{A},\mathbf{B}=\mathrm{Split}(\hat{\mathbf{X}}),
\end{equation}
\begin{equation}
    \mathbf{S}_{left},\mathbf{I}_{left}=\mathrm{Top-}r(\mathrm{Max}(\mathbf{A}\mathbf{B})),
\end{equation}
\begin{equation}
    \hat{\mathbf{A}},\hat{\mathbf{B}}=\mathrm{Fusion}(\mathbf{A},\mathbf{B},\mathbf{S}_{left},\mathbf{I}_{left}),
\end{equation}
\begin{equation}
    \mathbf{X}_m=[\hat{\mathbf{A}},\hat{\mathbf{B}}],
\end{equation}
where, $\mathbf{S}_{left}\in \mathbb{R}^{r\times 1}$ and $\mathbf{I}_{left}\in \mathbb{R}^{r\times 1}$ denote the similarity scores and indices of the left nodes of the $r$ edges, respectively. The operation $[\cdot]$ represents channel-wise concatenation, while $\mathrm{Max}(\cdot)$ selects the edge with the highest similarity for each token in the set $\mathbf{A}$. $\mathrm{Fusion}(\cdot)$ denotes the process of fusing the features of the nodes for each edge. $\mathbf{X}_m \in \mathbb{R}^{(n-r)\times C}$ is the merged output.

\begin{figure*}
\centering
\includegraphics[width=\textwidth]{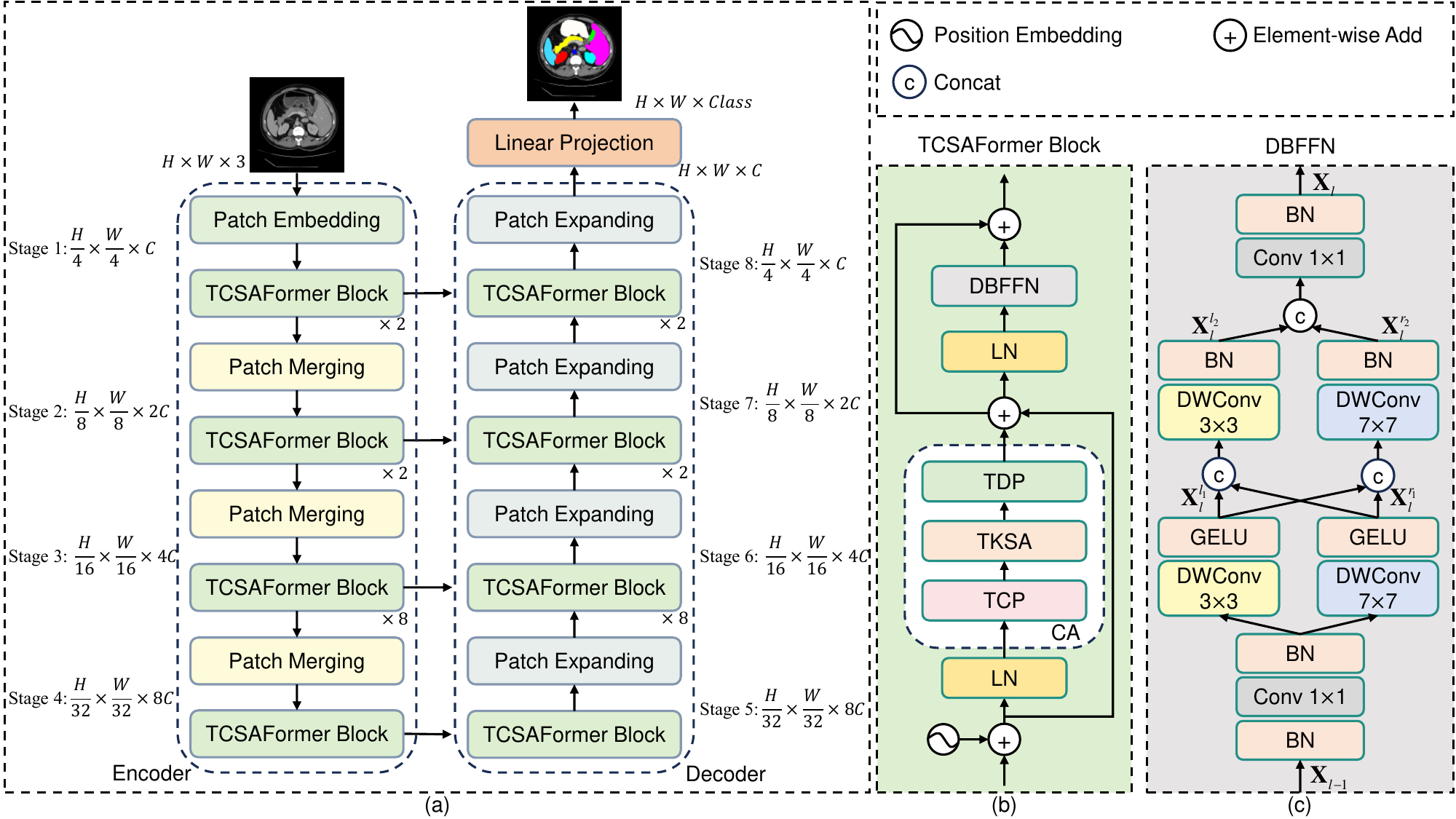}
\caption{(a) illustrates the overall architecture of TCSAFormer, which adopts a symmetric U-shaped encoder-decoder structure. (b) shows the detailed design of the TCSAFormer Block, consisting of the Compressed Attention (CA) and the Dual-Branch Feed-Forward Network (DBFFN). (c) presents the detailed design of the proposed DBFFN, which consists of primary depthwise convolutions with 3$\times$3 and 7$\times$7 kernels to effectively capture multiscale contextual information.}
\label{architecture}
\end{figure*}
\subsubsection{Top-k Sparse Attention}
To effectively enhance the computational efficiency of attention mechanisms while strengthening their focus on crucial features, we propose a content-aware dynamic Top-k sparse attention mechanism. This mechanism takes the output $\mathbf{X}_m$ from the token compression pipeline as input, and employs learnable parameter matrices $\{\mathbf{W}_q,\mathbf{W}_k,\mathbf{W}_v\}\in \mathbb{R}^{C \times C}$ to generate queries $\mathbf{Q}_m \in \mathbb{R}^{(n-r)\times C}$, keys $\mathbf{K}_m \in \mathbb{R}^{(n-r)\times C}$, and values $\mathbf{V}_m \in \mathbb{R}^{(n-r)\times C}$, respectively. 
\begin{equation}\mathbf{Q}_m=\mathbf{X}_m\mathbf{W}_q,\mathbf{K}_m=\mathbf{X}_m\mathbf{W}_k,\mathbf{V}_m=\mathbf{X}_m\mathbf{W}_v.\end{equation}
Then, a semantic relevance matrix $\mathbf{A}_m \in \mathbb{R}^{(n-r)\times (n-r)}$ is computed by taking the dot product between $\mathbf{Q}_m$ and $\mathbf{K}_m$. Following this, a Top-k selection procedure is applied to retain the $k$ most relevant key-value pairs for per query, producing a weight matrix $\mathbf{A}\in \mathbb{R}^{(n-r)\times k}$ and its corresponding index matrix $\mathbf{I}\in \mathbb{R}^{(n-r)\times k}$. Here, the value of $k$ is determined based on the length of the input sequence and the ratio $\lambda$.
\begin{equation}
    k=\lambda*(n-r),
\end{equation}
\begin{equation}
    \mathbf{A}_m=\mathbf{Q}_m(\mathbf{K}_m)^T,
\end{equation}
\begin{equation}
    \mathbf{A},\mathbf{I}=\mathrm{Top-}k(\mathbf{A}_m).
\end{equation}
Unlike the previous mask-based Top-$k$ attention, to prevent irrelevant tokens from participating in the computation, we gather $\mathbf{V}_m$ based on the indices in $\mathbf{I}$, and then perform weighted fusion using $\mathbf{A}$. The output of the attention layer is denoted by $\mathbf{X}_a \in \mathbb{R}^{(n-r) \times C}$.
\begin{equation}
    \mathbf{X}_a=\mathrm{Softmax}(\mathbf{A})\mathrm{Gather}(\mathbf{V}_m,\mathbf{I}).
\end{equation}
\subsubsection{Token Decompression Pipeline}
To effectively restore the spatial positional information of merged and pruned tokens,  we construct a Token Decompression Pipeline that integrates token unmerging and shortcut connection. First, the token unmerging operation, serving as the inverse process of token merging, also consists of three key steps: 
\begin{enumerate}
    \item Splitting $\mathbf{X}_a$ into set $\mathbf{A}^{\prime}$ and set $\mathbf{B}^{\prime}$ based on their spatial distribution.
    \item Filling the vacant positions in $\mathbf{A}^{\prime}$ using $\mathbf{B}^{\prime}$ and  $\mathbf{I}_{left}$, resulting in $\mathbf{A}^{\prime \prime}$.
    \item Constructing $\mathbf{A}^{\prime \prime}$ and $\mathbf{B}^{\prime}$ into the complete feature set $\mathbf{X}_d\in \mathbb{R}^{n \times C}$.
\end{enumerate}
The decompression procedure is illustrated in Fig. \ref{CAM} (d), and can be formulated as:
\begin{equation}
    \mathbf{A}^{\prime},\mathbf{B}^{\prime}=\mathrm{Split}(\mathbf{X}_a),
\end{equation}
\begin{equation}
        \mathbf{A}^{\prime \prime}=\mathrm{Fill}(\mathbf{B}^{\prime},\mathbf{I}_{left}),
\end{equation}
\begin{equation}
    \mathbf{X}_{d}=\mathrm{Construct}(\mathbf{A}^{\prime \prime},\mathbf{B}^{\prime}),
\end{equation}
where, the shapes of $\mathbf{A}^{\prime}, \mathbf{B}^{\prime}, \mathbf{A}^{\prime \prime}$ are all $\frac{n}{2} \times C$.

Second, for the shortcut connection, we restore the original spatial positions of tokens in $\mathbf{X}_d$ through $\mathbf{M}$, then execute element-wise addition with $\mathbf{X}_p$ to obtain the output $\mathbf{O}\in \mathbb{R}^{HW \times C}$.
\begin{equation}
        \mathbf{O}=\mathbf{X}_p+\mathrm{Gather}(\mathbf{X}_{d},\mathbf{M}).
\end{equation}

\subsection{Dual-Branch Feed-Forward Network}
Unlike the linear projection-based feed-forward networks, we propose DBFFN to better model local context and fuse multiscale features. As shown in Fig. \ref{architecture} (c), DBFFN first performs cross-channel information interaction of the input features using a 1$ \times $1 convolution. It then constructs two lightweight depthwise convolutional branches to extract multiscale imformation: the left branch uses a 3$ \times $3 kernel to capture fine-grained local spacial information, while the right branch uses a 7$ \times $7 kernel to expand the receptive field and extract coarse-grained regional features. Finally, another 1$ \times $1 convolution is used to fuse the information from the two branches. Additionally, the architecture incorporates the GELU activation function to enhance non-linear modeling capabilities and uses Batch Normalization (BN) to accelerate the model’s convergence process.
 Given an input feature map $\mathbf{X}_{l-1} \in \mathbb{R}^{HW \times C}$, the entire feature fusion process in DBFFN is formulated as:
\begin{equation}
    {\mathbf{\hat{X}}_l}=\mathrm{BN}(\mathrm{Conv}_{1 \times 1}(\mathrm{BN}(\mathbf{X}_{l-1}))),
\end{equation}
\begin{equation}
    \mathbf{X}_l^{l_1}=\sigma(\mathrm{DWConv}_{3\times 3}(\mathbf{\hat{X}}_l)), 
\end{equation}
\begin{equation}
    \mathbf{X}_l^{r_1}=\sigma(\mathrm{DWConv}_{7\times 7}(\hat{\mathbf{X}_l})),
\end{equation}
\begin{equation}
\mathbf{X}_l^{l_2}=\mathrm{BN}(\mathrm{DWConv}_{3\times 3}([\mathbf{X}_l^{l_1}, \mathbf{X}_l^{r_1}]
)), 
\end{equation}
\begin{equation}
    \mathbf{X}_l^{r_2}=\mathrm{BN}(\mathrm{DWConv}_{7\times 7}([\mathbf{X}_l^{l_1}, \mathbf{X}_l^{r_1}
]
)),
\end{equation}
\begin{equation}
    \mathbf{X}_l=\mathrm{BN}(\mathrm{Conv}_{1 \times 1}(
    [\mathbf{X}_l^{l_2}, \mathbf{X}_l^{r_2}
])),
\end{equation}
where, $\sigma(\cdot)$ denotes GELU activation, $[\cdot]$ is the channel-wise concatenation, $\{\hat{\mathbf{X}_l}, \mathbf{X}_l^{l_1}, \mathbf{X}_l^{r_1}\}\in \mathbb{R}^{HW\times 2C}$ and $\{\mathbf{X}_l^{l_2}, \mathbf{X}_l^{r_2}\}\in \mathbb{R}^{HW\times C}$ are intermediate states, and $\mathbf{X}_l \in \mathbb{R}^{HW \times C}$ is the final output.

\subsection{Transformer Block}
Based on CA and DBFFN, we construct an efficient transformer block named TCSAFormer block, with the detailed architecture illustrated in Fig. \ref{architecture} (b). The block consists of three components: First, a 3$\times$3 depthwise convolution is employed to encode relative positional information. Subsequently, the CA implements an efficient multi-head attention mechanism to model global contextual relationships. Finally, the DBFFN adopts a dual-branch structure for multiscale feature extraction, thereby enhancing the feature representation capability. Additionally, layer normalization and residual connections are incorporated to ensure training stability.

\subsection{TCSAFormer 
 Architecture}
 Fig. \ref{architecture} (a) illustrates the network architecture of our proposed TCSAFormer, which consists of an encoder, a decoder, skip connections, and a linear projection layer.

The encoder consists of four stages. Specifically, in the stage 1, an input image of size $H \times W \times 3$ is processed through a Patch Embedding layer, which performs overlapping patch partitioning and linear embedding, resulting in a feature map of size $\frac{H}{4} \times \frac{W}{4} \times C$. This feature map is then passed through $N_1$ TCSAFormer blocks for feature extraction. The subsequent three stages each comprise a Patch Merging layer followed by $N_i$ consecutive Transformer blocks, where $i$ indicates the current stage. Patch Merging layer performs 2$\times$ downsampling and increases the channel dimensions by 2$\times$.
Therefore, the three stages respectively produce feature maps of sizes $\frac{H}{8} \times \frac{W}{8} \times 2C$, $\frac{H}{16} \times \frac{W}{16} \times 4C$, and $\frac{H}{32} \times \frac{W}{32} \times 8C$.
The structure of the decoder is symmetric to the encoder, with the key difference being the use of Patch Expanding instead of Patch Embedding and Patch Merging. Patch Expanding performs 2$\times$ upsampling while reducing channel dimensions by half. The exception is the final Patch Expanding layer, which applies 4$\times$ upsampling to restore the original image resolution.
To address potential information loss in the feature map during the downsampling process, we introduce skip connections between the encoder and decoder. Moreover, a linear projection layer is used to convert the channel dimensions to the number of output classes, generating the final segmentation prediction.
In our experiments, the number of Transformer blocks $N_i$ at each stage $i$ is set to [2, 2, 8, 1, 1, 8, 2, 2].
\section{Experiments}
\label{experiment}
In this section, we first introduce the dataset used. Then, we provide a detailed description of the Experimental settings and present the quantitative and qualitative results for each dataset. Finally, we conduct a series of ablation studies to validate the effectiveness of each component involved in TCSAFormer.
\begin{table*}
\centering
\small
\caption{Quantitative results on DSC, and HD of our approach against other state-of-the-art methods for medical image segmentation on Synapse multi-organ segmentation dataset. Only DSC is exclusively used for the evaluation of individual organ. The symbol $\uparrow$ indicates the larger the better. The symbol $\downarrow$ indicates the smaller the better. The best result is in \textbf{Blod}, and the second best is \underline{underlined}.}
\resizebox{1.0\linewidth}{!}{
\begin{tabular}{l|cc|ccccccccc}
\toprule
Methods & DSC (\%) $\uparrow$  & HD (mm) $\downarrow$ & Aorta & Gallbladder & Kidney(L) &Kidney(R) &Liver & Pancreas & Spleen & Stomach\\
\midrule 
TransUNet \cite{chen2021transunet}  & 77.48   &31.69 &87.23 &63.13 &81.87 &77.02 &94.08 &55.86 &85.08 &75.62 
\\
Swin-Unet \cite{cao2022swin}  & 79.13  & 21.55 & 85.47 & 66.53  & 83.28 & 79.61  & 94.29 & 56.58 & 90.66  & 76.60
\\
HiFormer \cite{heidari2023hiformer}  & 80.39  & \textbf{14.70} & 86.21 &65.69  & 85.23 &79.77  &94.61 & 59.52 &90.99 & 81.08
\\
MISSFormer \cite{huang2022missformer}  &81.96  & 18.20 & 86.99 &68.65  &85.21& \underline{82.00} &94.41 & 65.67 &\textbf{91.92} & 80.81
\\
MT-UNet \cite{wang2022mt-unet} & 78.59 & 26.59 & 87.92 & 64.99 & 81.47 & 77.29 & 93.06 & 59.46 & 87.75 & 76.81\\
MSSAANet \cite{zeng2023msaanet} & \underline{82.85} & 18.54 & \textbf{89.40} & \textbf{73.20} & 84.31 & 78.53 & \textbf{95.10} & \underline{68.85} & 91.60 & 81.78\\
BRAU-Net++ \cite{Lan2024BRAU-Net++} & 82.47 &19.07 & \underline{87.95} &69.10  &\textbf{87.13} &81.53 & 94.71  & 65.17 &\underline{91.89} & \underline{82.26}\\
\midrule 
TCSAFormer  & \textbf{83.16} & \underline{17.91} & 87.16 & \underline{71.25} & \underline{86.18} & \textbf{82.51} & \underline{94.95} & \textbf{69.02}  & 91.09 & \textbf{83.12}\\
\bottomrule
\end{tabular} }
\label{tab1:synapse}
\end{table*}
\begin{figure*}
\centering
\includegraphics[width=\textwidth]{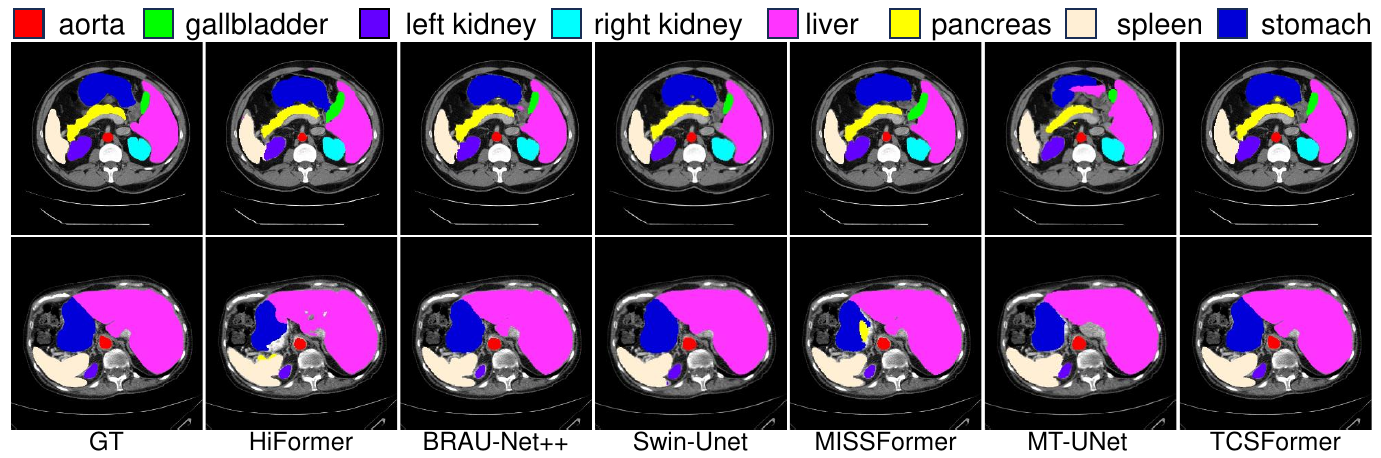}
\caption{Visualization of segmentation results comparing the proposed TCSAFormer with state-of-the-art medical image segmentation methods on the Synapse dataset. As demonstrated, TCSAFormer accurately delineates organ boundaries while effectively preserving internal structural details of organs, presenting more precise and comprehensive segmentation results.}
\label{Synapse}
\end{figure*}

\subsection{Datasets}
\subsubsection{Synapse}
The Synapse dataset is a multi-organ segmentation dataset consisting of 30 abdominal CT scans, comprising 3,779 axial clinical abdominal CT images. Following \cite{chen2021transunet,cao2022swin,huang2022missformer}, 18 scans are used for training and the remaining 12 for testing. To quantitatively evaluate the performance of our method, we adopt the Dice Similarity Coefficient (DSC) and the 95\% Hausdorff Distance (HD) as evaluation metrics.
\subsubsection{ISIC-2018}
 This dataset is widely recognized for benchmarking methods in skin lesion segmentation, contains 2,594 clinical dermoscopy images. In our experiments, the dataset is randomly split into 1,868 training images, 467 validation images, and 259 test images. We employ Mean Intersection over
Union (mIoU), DSC, Accuracy, Precision, and Recall as evaluation metrics.
\subsubsection{CVC-ClinicDB}
The CVC-ClinicDB dataset is commonly used for polyp image segmentation, consisting of 612 images. In our study, the dataset is partitioned into 490 images for training, 61 images for validation, and 61 images for testing, providing a balanced distribution for model evaluation. The evaluation metrics for this dataset are consistent with ISIC-2018.

\subsection{Experimental settings}
We train TCSAFormer on an NVIDIA 3090 graphics card with 24GB memory and use PyTorch 2.0. The encoder of TCSAFormer is initialized with pre-trained weights from the ImageNet dataset. To improve the generalization ability of the model, we employ a series of data augmentation techniques, including random flipping, rotation, and Cutout. The specific settings for different datasets are as follows:
\begin{itemize}
    \item \textbf{ISIC-2018} and \textbf{CVC-ClinicDB}: resolution = 256$\times$256; lr = 1$e$-4; lr-scheduler = Cosine Annealing; optimizer = Adam; epochs = 200; batch size = 16, loss = CE (Cross-Entropy Loss).
    \item \textbf{Synapse}: resolution = 224$\times$224; lr = 1$e$-4; lr-scheduler = Cosine Annealing; optimizer = AdamW (weight decay = 1$e$-4); epochs = 400; batch size = 24, loss = 0.5 $\times$ CE (Cross-Entropy Loss) + 0.5 $\times$ DICE (Dice Loss).
\end{itemize}
Moreover, the parameter settings of TCSAFormer are as follows: the dimension of each attention head is set to 32. Based on a grid search, the channel dimensions of the encoder stages are set to [64, 128, 256, 512], with corresponding token pruning ratios of [0.5, 0.4, 0.3, 0.1] and token merging ratios of [0.3, 0.2, 0.1, 0.1]. The hyperparameter $\lambda$ is set to 1/8. The decoder adopts the same configuration as the encoder.

\subsection{Results}
\subsubsection{Comparison on Synapse}
Table \ref{tab1:synapse} presents the comparative results between TCSAFormer and state-of-the-art medical image segmentation methods on the Synapse dataset. The results demonstrate that TCSAFormer achieves the best average Dice coefficient of 83.16\%, indicating high consistency between its segmentation predictions and ground truth annotations. In terms of the average HD, TCSAFormer attains a second best score of 17.91 mm. It suggests that there is still room for improvement in accurately capturing boundary information. Furthermore, regarding single organ segmentation, TCSAFormer achieves the best or second best results in terms of the Dice metric for six of the eight organs.

Fig. \ref{Synapse} illustrates the visualization results of TCSAFormer compared with other methods in Table \ref{tab1:synapse} on the Synapse dataset, intuitively demonstrating the superiority of our method. From these visualization results, it can be clearly observed that TCSAFormer can precisely outline organ boundaries and accurately preserve internal structural details of organs, while other methods generally suffer from problems such as unclear internal details, blurred organ boundaries, and incorrect category recognition.

\begin{table*}[htpb]
\centering
\caption{Quantitative results of different methods for Medical Image Segmentation on ISIC-2018 Segmentation and CVC-ClinicDB datasets.}
\resizebox{1.0\linewidth}{!}
{
    \begin{tabular}{l|ccccc|ccccc}
        \toprule
        \multirow{2}{*}[-0.5ex]{Methods} & \multicolumn{5}{c|}{ISIC-2018} & \multicolumn{5}{c}{CVC-ClinicDB}\\
        \cmidrule{2-11}
         & mIoU (\%) & DSC (\%) & Acc (\%) & Prec (\%) & Recall (\%) & mIoU (\%) & DSC (\%) & Acc (\%) & Prec (\%) & Recall (\%) \\
        \midrule
       MedT \cite{valanarasu2021medt} & 81.43 & 86.92 & 95.10 & 90.56 & 89.93 & 81.47 & 86.97 & 98.44 & 89.35 & 90.04  \\
       TransUNet \cite{chen2021transunet} & 77.05 & 84.97 & 94.56 & 84.77 & 89.85 & 79.95 & 86.70 & 98.25 & 87.63 & 87.34 \\
       Swin-Unet \cite{cao2022swin} & 81.87 & 87.43 & 95.44 & 90.97 & 91.28 & 84.85 & 88.21 & 98.72 & 90.52 & 91.13 \\
       HiFormer \cite{heidari2023hiformer} & 81.54 & 88.44 & 94.68 & 90.23 & 90.82 & 85.74 & 90.48 & 98.62 & 91.56 & 90.37\\ 
       MISSFormer \cite{huang2022missformer} & 83.86 & 90.02 & 94.67 & \underline{91.65} & 92.04 & 86.02 & 91.46 & 98.74 & 93.06 & 90.97\\
       BRAU-Net++ \cite{Lan2024BRAU-Net++} & \underline{84.01} & \underline{90.10} & \underline{95.61} & 91.18 & \underline{92.24} & \underline{88.17} & \underline{92.94} & \underline{98.83} & \underline{93.84} & \underline{93.06} \\
       \midrule
       TCSAFormer & \textbf{84.12} & \textbf{90.23} & \textbf{95.65} & \textbf{91.74} & \textbf{92.33} & \textbf{90.54} & \textbf{94.90} & \textbf{99.11} & \textbf{95.54} & \textbf{94.63}\\
        \bottomrule
    \end{tabular}
}
\label{tab2:isic & cvc}
\end{table*}
\begin{figure*}
\centering
\includegraphics[width=\textwidth]{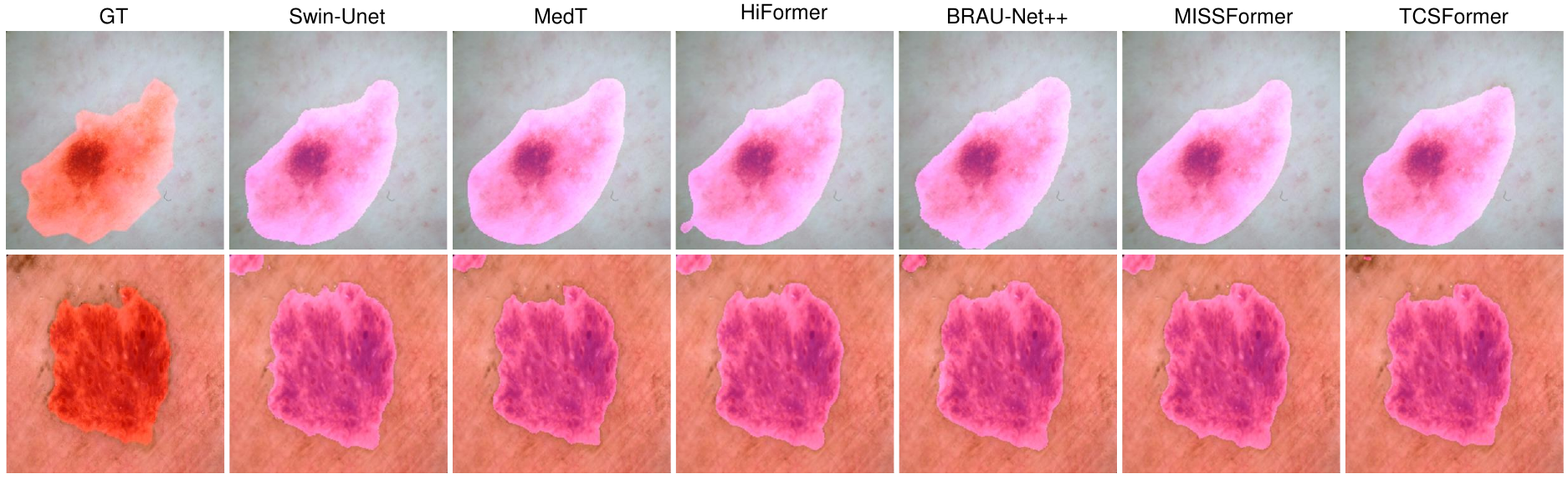}
\caption{Visualization of segmentation results comparing the proposed TCSAFormer with state-of-the-art medical image segmentation methods on the ISIC-2018 dataset.}
\label{ISIC}
\end{figure*}
\begin{figure*}
\centering
\includegraphics[width=\textwidth]{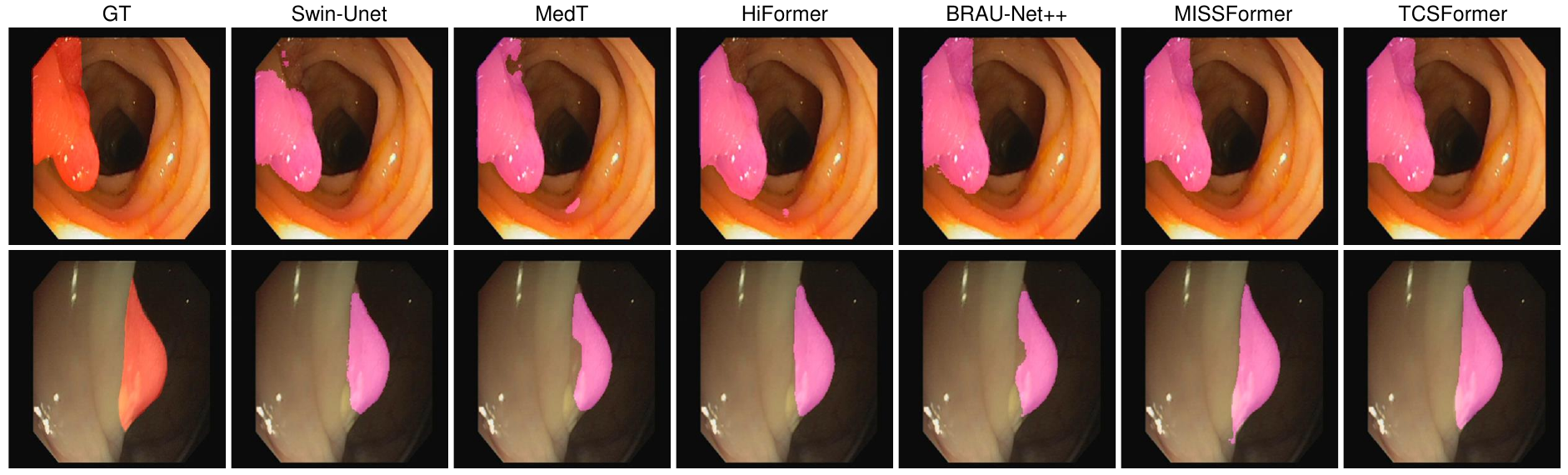}
\caption{Visualization of segmentation results comparing the proposed TCSAFormer with state-of-the-art medical image segmentation methods on the CVC-ClinicDB dataset.}
\label{CVC}
\end{figure*}
\subsubsection{Comparison on ISIC-2018}
To evaluate the performance of our method, we conduct a quantitative comparison between TCSAFormer and other state-of-the-art segmentation methods on the ISIC-2018 dataset. The evaluation results are shown in Table \ref{tab2:isic & cvc}. The results indicate that TCSAFormer achieves the best performance across five key metrics, significantly outperforming the other compared methods. Although BRAU-Net++ and MISSFormer achieve relatively good segmentation performance, there is still a certain gap compared to our method. Our analysis suggests that the superior performance of TCSAFormer mainly benefits from the effective integration between the innovative TCP and TKSA, which enables the model to capture and focus on the most relevant features, thereby greatly enhancing its feature representation capability. Fig. \ref{ISIC} presents the visual segmentation results of TCSAFormer and other methods on the ISIC-2018 dataset. These results clearly demonstrate the superior segmentation performance of TCSAFormer.

\subsubsection{Comparison on CVC-ClinicDB}
Table \ref{tab2:isic & cvc} presents the qualitative comparison between TCSAFormer and other state-of-the-art segmentation methods on the CVC-ClinicDB dataset. TCSAFormer achieves outstanding performance across five metrics, including an mIoU of 90.54\%, a DSC of 94.90\%, an accuracy of 99.11\%, a precision of 95.54\%, and a recall of 94.63\%, significantly outperforming existing methods. These results demonstrate that the segmentation outputs produced by TCSAFormer are highly consistent with the ground truth, enabling accurate pixel-level classification. Fig. \ref{CVC} further illustrates the visual segmentation results of various medical image segmentation methods on the CVC-ClinicDB dataset.
\subsection{Ablation Studies}
In this section, we conduct a series of ablation studies on the Synapse dataset to validate the effectiveness of each component in the proposed TCSAFormer. Specifically, we perform an attention visualization analysis of the CA and quantitatively evaluate the performance of TCP and DBFFN.

\subsubsection{Visualization of Attention Map}
\begin{figure}
\centering
\includegraphics[width=\columnwidth]{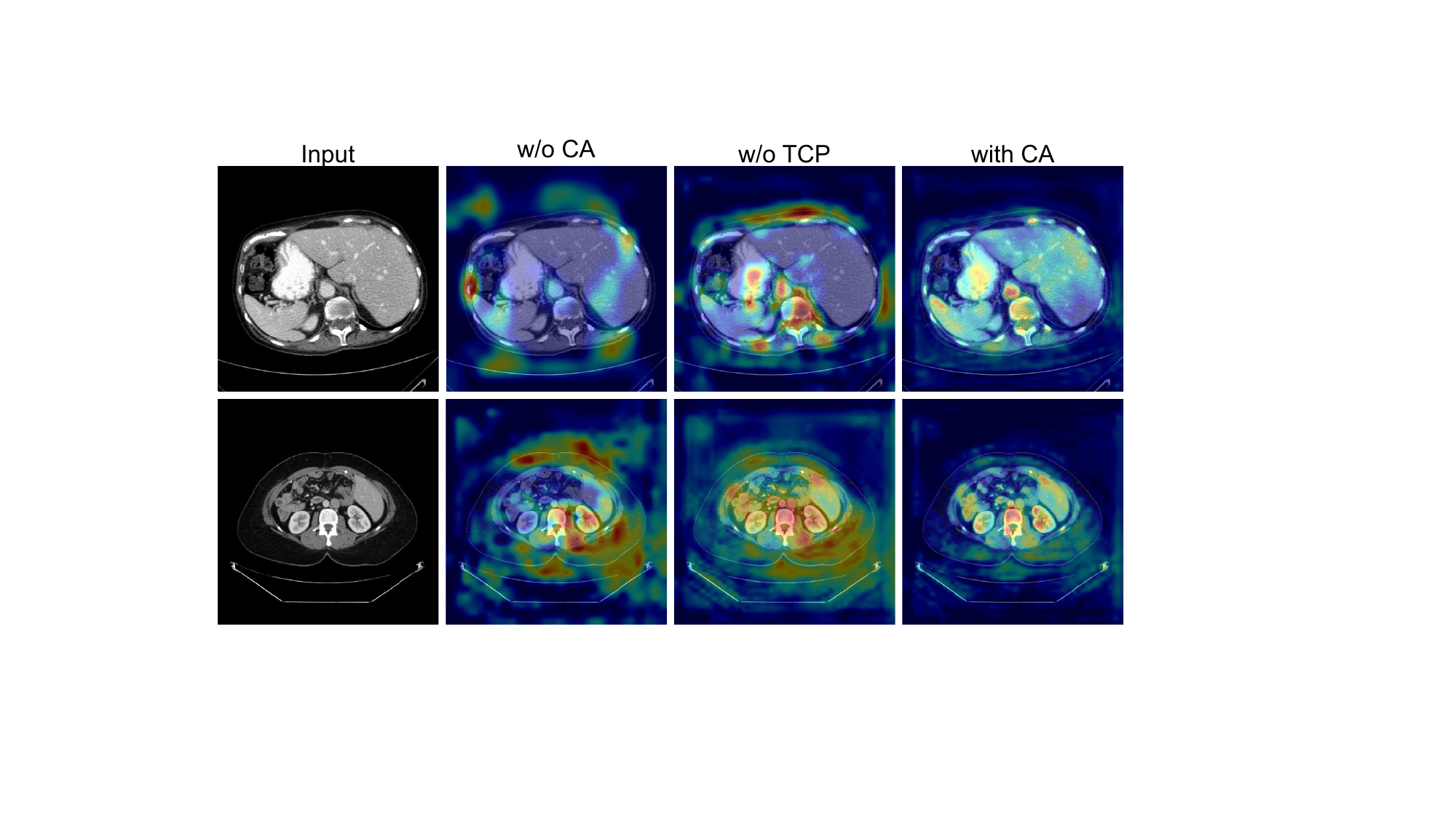}
\caption{An ablation analysis of the components within the CA module using Grad-CAM visualization. From left to right, the four images show: the original input image, the attention map without the CA, the attention map without TCP but with TKSA, and the attention map with the full CA.}
\label{grad_cam}
\end{figure}

We employ Grad-CAM \cite{selvaraju2017grad} to visualize the attention of the CA on the Synapse dataset, providing an intuitive demonstration of the effectiveness of the proposed CA. The results are shown in Fig. \ref{grad_cam}. It is evident that, without CA, the model allocates most of its attention to the background, failing to focus effectively on the target regions. After introducing TKSA, the model better directs attention to the organ regions, although the background still receives some focus. Finally, when the  CA is used, the model achieves significantly improved attention localization on targets.
In summary, both TKSA and TCP effectively capture critical features. Building upon this, the proposed CA integrates their strengths to progressively refine the attention distribution, enabling more precise focus on relevant regions.
\subsubsection{Effectiveness of TCP}
Under the TCSAFormer architecture, we evaluate the effectiveness of the proposed TCP by comparing it with various token compression methods. The baseline model is TCSAFormer without token compression, and all token compression methods use the same number of compression tokens as TCP to ensure a fair comparison. As shown in Table \ref{ablation:token compression}, the proposed TCP achieves the best segmentation performance while reducing FLOPs by 34.7\%, providing the optimal trade-off between efficiency and accuracy.
Similarly, DynamicViT also reduces FLOPs by 34.5\%, but excessive token pruning results in the loss of critical information, leading to reduced model performance.  Moreover, AdaViT and PITOME show more modest reductions in FLOPs, as they rely on computing token similarities for pruning or merging, which is less efficient. Overall, the experimental results demonstrate that TCP combines token pruning and token merging to effectively reduce model complexity while preserving key information, thereby validating the rationale behind our design.

\subsubsection{Effectiveness of DBFFN}
This work conducts a series of ablation experiments under the TCSAFormer architecture to evaluate the effectiveness of the proposed DBFFN by replacing different FFN modules. The standard TCSAFormer with MLP-FFN is used as the baseline, and comparative models are constructed by integrating various FFN variants, including Con-FFN, Ghost FFN, ReMix-FFN, CGLU, and DBFFN. Detailed experimental results are presented in Table \ref{ablation:ffn}. The analysis shows that, compared to the baseline, these FFN variants enhance feature representation by incorporating components such as convolution and residual connections, leading to improved segmentation performance. Among them, only Ghost-FFN reduces model complexity through an innovative lightweight expand-reduce strategy, other methods increase computational overhead due to extra convolutional operations.
In contrast, our proposed DBFFN employ dual-branch architecture fusing multiscale feature, and replace conventional MLP with convolutional structures, thus achieving best segmentation performance while decreasing computation overhead. This design provides a new technical perspective for the lightweight of medical image segmentation networks.

\begin{table}[]
    \centering
    \caption{Ablation Study on the Effectiveness of TCP.}
    \resizebox{1.0\linewidth}{!}{
        \begin{tabular}{cc|cc|c}
            \toprule
             Architecture &  Token Compression & DSC (\%) & HD (mm) & FLOPs (G)\\
             \midrule
             TCSAFormer & - & 78.82 & 24.89 & 7.96\\
             TCSAFormer & DynamicViT \cite{rao2021dynamicvit} & 78.55 & 25.61 & 5.21 \color{blue}{(-34.5\%)}\\
             TCSAFormer & AdaViT \cite{meng2022adavit} & 78.98 & 24.54 & 7.76 \color{blue}{(-2.5\%)}\\
             TCSAFormer & ToMe \cite{bolya2023tome} & 79.05 & 24.66 & 6.14 \color{blue}{(-22.9\%)}\\
             TCSAFormer & PITOME \cite{tran2024pitome} & 79.22 & 24.35 & 8.44 \color{blue}{(-6.0\%)}\\
             TCSAFormer & TCP (Ours) & \textbf{79.31} & \textbf{23.51} & \textbf{5.20} \color{blue}{(-34.7\%)}\\
             \bottomrule
        \end{tabular}
    }
    \label{ablation:token compression}
\end{table}

\begin{table}[]
    \centering
    \caption{Ablation Study on the Effectiveness of DBFFN.}
    \resizebox{1.0\linewidth}{!}{
        \begin{tabular}{cc|cc|c}
            \toprule
             Architecture &  FFN & DSC (\%) & HD (mm) & FLOPs (G)\\
             \midrule
             TCSAFormer & MLP-FFN($e$=4) \cite{dosovitskiy2020vit} & 77.71 & 28.32 & 6.00\\
             TCSAFormer & Con-FFN \cite{wang2022pvtv2,xie2021segformer} & 77.89 & 28.16 & 6.07 \color{blue}{(+1.2\%)}\\
             TCSAFormer & Ghost FFN \cite{cao2023ghostvit} & 78.21 & 25.86 & 5.73 \color{blue}{(-4.5\%)}\\
             TCSAFormer & ReMix-FFN \cite{huang2022missformer} & 78.65 & 25.46 & 6.20 \color{blue}{(+3.3\%)}\\
             TCSAFormer & CGLU \cite{li2024dmsa}& 78.39 & 26.12 &   6.11 \color{blue}{(+1.8\%)}\\
             TCSAFormer & DBFFN (Ours) & \textbf{79.31} & \textbf{23.51} & \textbf{5.20} \color{blue}{(-13.3\%)}\\
             \bottomrule
        \end{tabular}
        }
    
    \label{ablation:ffn}
\end{table}
\section{Conclusion}
\label{conclusion}
In this paper, we propose an efficient medical image segmentation network named TCSAFormer, which is composed of the efficient attention module CA and the representation-enhancing DBFFN. CA incorporates both token compression and sparse attention mechanism to reduce the computational complexity of self-attention and effectively model global dependencies among tokens. DBFFN adopts a dual-branch convolutional structure to address the limitations of existing methods in capturing local spatial details and multiscale features, thereby substantially enhancing the model's representation capability. Benefiting from these designs, TCSAFormer achieves an excellent balance between computational efficiency and segmentation accuracy, and demonstrates superior performance on three publicly available medical image segmentation datasets.

Although TCSAFormer demonstrates superior segmentation performance, there is still room for improvement. Currently, the pruning thresholds and merging ratios at each stage are manually set as hyperparameters, which limits adaptability and generalizability. Therefore, future work will focus on enabling the model to automatically learn optimal token compression ratios during training, aiming to further enhance the performance of medical image segmentation.

\bibliographystyle{cas-model2-names}

\bibliography{cas-refs}

\end{document}